\pdfoutput=1

\documentclass[11pt]{article}

\usepackage[preprint]{acl}

\usepackage{times}
\usepackage{latexsym}

\usepackage[T1]{fontenc}

\usepackage[utf8]{inputenc}

\usepackage{microtype}

\usepackage{inconsolata}

\usepackage{graphicx}
\usepackage{amsmath} 
\usepackage{forest}
\usepackage{array}
\usepackage{multirow}
\usepackage{booktabs}
\usepackage{color}
\usepackage{amssymb} 
\title{A Survey on Data Contamination for Large Language Models}

\author{Yuxing Cheng$^{1}$ \quad  Yi Chang$^{2,3,4}$ \quad Yuan Wu$^{2}$\footnotemark[1] \\
        $^{1}$College of Software, Jilin University \\ 
        $^{2}$School of Artificial Intelligence, Jilin University \\
        $^{3}$Engineering Research Center of Knowledge-Driven Human-Machine Intelligence, MOE, China \\
        $^{4}$International Center of Future Science, Jilin University\\
        chengyx0121@mails.jlu.edu.cn, yichang@jlu.edu.cn, yuanwu@jlu.edu.cn \\   
}

\begin{document}

\maketitle
\renewcommand{\thefootnote}{\fnsymbol{footnote}}
\footnotetext[1]{Corresponding authors}

\begin{abstract}
Recent advancements in Large Language Models (LLMs) have demonstrated significant progress in various areas, such as text generation and code synthesis. However, the reliability of performance evaluation has come under scrutiny due to data contamination—the unintended overlap between training and test datasets. 
This overlap has the potential to artificially inflate model performance, as LLMs are typically trained on extensive datasets scraped from publicly available sources. These datasets often inadvertently overlap with the benchmarks used for evaluation, leading to an overestimation of the models' true generalization capabilities.
In this paper,
we first examine the definition and impacts of data contamination.  
Secondly, we review methods for contamination-free evaluation, focusing on three strategies: data updating-based methods, data rewriting-based methods, and prevention-based methods. Specifically, we highlight dynamic benchmarks and LLM-driven evaluation methods. Finally, we categorize contamination detecting methods based on model information dependency: White-Box, Gray-Box, and Black-Box detection approaches. Our survey highlights the requirements for more rigorous evaluation protocols and proposes future directions for addressing data contamination challenges.
\end{abstract}

\section{Introduction}
\label{sec:introduction}

Recent breakthroughs in Large Language Models (LLMs) have demonstrated remarkable capabilities in text generation, code synthesis, and mathematical reasoning~\cite{zhao2023survey,openai2024gpt4technicalreport,deepseekai2025deepseekr1incentivizingreasoningcapability}.
However, the reliability of LLM evaluation is increasingly questioned due to data contamination-the unintended overlap between training and test data sets \cite{balloccu-etal-2024-leak,chang2024survey}. This is especially problematic as LLMs use large web-scraped datasets that are prone to overlap with testing benchmarks. \cite{xu2024benchmarkingbenchmarkleakagelarge} analyzed 31 LLMs in the context of mathematical reasoning, uncovering widespread data contamination. 
LLMs are known to memorize portions of their training data, and under certain prompts, they can reproduce this data verbatim \cite{carlini2022quantifying}. As highlighted by \cite{sainz-etal-2023-nlp}, a critical consequence of data contamination is that scientific studies relying on contaminated LLMs may produce erroneous conclusions, potentially invalidating valid hypotheses. 
To underscore the importance of addressing data contamination in both the development and evaluation of LLMs, we present a comprehensive review of data contamination.

In section \ref{sec:what}, we define data contamination as the inclusion of data from the testing set during the pre-training phase, which artificially inflates model performance. Recent studies extend this definition along two dimensions: phase-based contamination in LLMs' lifecycle and benchmark-based contamination in LLMs' evaluation. For phase-based analysis, contamination mechanisms include pre-training phase leakage, fine-tuning biases, cross-modal leakage \cite{yao-etal-2024-data}, and indirect human interactions \cite{palavalli-etal-2024-taxonomy}. 
Meanwhile, we divide benchmark-based contamination into four types: Text contamination, Text-label contamination, Augmentation-based contamination, and Benchmark-level contamination. We provide a fine-grained analysis of its impacts, including Evidence collection, Non-Contamination scenarios, Quantification of contamination, and Characteristics of data contamination.

In section \ref{sec:where}, we discuss how to achieve contamination-free evaluation. For static benchmarks, current research focuses on three key contamination-free strategies: automatically updating datasets using the most recent data, rewriting existing data, and implementing proactive risk prevention mechanisms. Meanwhile, dynamic evaluation frameworks\cite{zhu2024dyvaldynamicevaluationlarge,lei-etal-2024-s3eval,zhang2024darg,ying2024automating} generate test samples using techniques like combinatorial optimization, graph-based reasoning, and controlled randomization, creating an evolving evaluation system. Additionally, the LLM-as-a-judge paradigm\cite{bai2024benchmarking} turns LLMs into meta-evaluators, enabling intelligent assessments independent of static benchmarks.

In section \ref{sec:how}, we explore methodologies for detecting data contamination in LLMs. We categorize data contamination detection approaches into three distinct paradigms: white-box detection, which relies on full access to model architectures or training data to achieve high precision, employing techniques such as N-gram overlap \cite{brown2020language} or embedding similarity \cite{reimers2019sentence}; gray-box detection, which leverages partial model information, such as token probabilities, to identify contamination; and black-box detection, which operates without access to internal model details, relying instead on heuristic rules (the details are outlined in Appendix \ref{sec:definition of assumptions}). Together, these approaches illustrate the evolving and multifaceted landscape of data contamination detection methods, each offering unique advantages and challenges.

\begin{figure*}[h]
\centering
\begin{forest}
  for tree={
    grow=east,
    reversed=true,
    anchor=base west,
    parent anchor=east,
    child anchor=west,
    base=left,
    font=\small,
    rectangle,
    draw,
    rounded corners,
    align=left,
    minimum width=2.5em,
    inner xsep=4pt,
    inner ysep=1pt,
  },
  where level=1{
  fill=blue!10}{},
  where level=2{
  font=\footnotesize, fill=pink!30}{},
  where level=3{font=\footnotesize, yshift=0.26pt, fill=yellow!20, 
  }{},
  [Data Contamination, fill=green!20
    [What is data \\ contamination \\ (Sec.~\ref{sec:what}), text width=4.4em
      [{Contamination\\Types}
        [{
        Phase-based Contamination:\\ \citeauthor{sainz-etal-2023-nlp},\citeauthor{balloccu-etal-2024-leak},  
            \citeauthor{song2024textimagesleakedsystematic}
        \\Benchmark-based Contamination:\\
    \citeauthor{jiang2024investigatingdatacontaminationpretraining} 
            \citeauthor{palavalli-etal-2024-taxonomy}  
        }]
      ]
      [Impacts
        [{Evidence Collection:\\ 
            \citeauthor{li2024task}, \citeauthor{riddell2024quantifyingcontaminationevaluatingcode}
            \citeauthor{liu-etal-2024-evaluating},
            \citeauthor{sainz-etal-2023-nlp}
        \\Non-Contamination Scenarios:\\
        \citeauthor{dekoninck2024constat} 
       \citeauthor{palavalli-etal-2024-taxonomy}
        \\Quantifying Methods:\\
        \citeauthor{brown2020language, touvron2023llama2, singh2024evaluation}  
    \citeauthor{li2023estimatingcontaminationperplexityquantifying}  
     \\Characteristics of Data Contamination:\\
        \citeauthor{villalobos2024rundatalimitsllm},
        \citeauthor{deng2023investigating},
        \citeauthor{kocyigit2025overestimationllmevaluationcontrolled},\citeauthor{riddell2024quantifyingcontaminationevaluatingcode},
        \citeauthor{balloccu-etal-2024-leak}\\
        \citeauthor{palavalli-etal-2024-taxonomy},\citeauthor{golchin2023time},\citeauthor{jiang2024investigatingdatacontaminationpretraining},\citeauthor{yao-etal-2024-data},\citeauthor{zhang-etal-2024-pacost}
        }]
      ]
    ]
    [How to \\ avoid data \\ contamination \\ (Sec.~\ref{sec:where}), text width=4.8em
      [{Benchmark \\Contamination-free}
        [{Data Updating-based Methods:\\
        \citeauthor{li2024latestevaladdressingdatacontamination}, \citeauthor{white2024livebench}, \citeauthor{jain2024livecodebench},
        \citeauthor{li2024evocodebench},
        \citeauthor{fan2024nphardeval4vdynamicreasoningbenchmark}   
        \\Data Rewriting-based Methods:\\
        \citeauthor{zhu-etal-2024-clean}, \citeauthor{zhao2024mmlu}, 
        \citeauthor{li-etal-2023-cleva},\citeauthor{ying2024automating}
        \\Prevention-based Methods:\\
        \citeauthor{zhu2024inference}, \citeauthor{li2024c2levacomprehensivecontaminationfreelanguage}
         }]
      ]
      [{Dynamic Evaluation}
        [{\citeauthor{zhu2024dyvaldynamicevaluationlarge}, \citeauthor{lei-etal-2024-s3eval}, \citeauthor{zhang2024darg}\\
        \citeauthor{srivastava2024functional}, \citeauthor{qian2024varbench}, \citeauthor{wang2024benchmark}
        }]
      ]
      [{LLM-as-a-Judge}
        [{\citeauthor{bai2024benchmarking}, \citeauthor{yu-etal-2024-kieval}, \citeauthor{li2024treeevalbenchmarkfreeevaluationlarge},\citeauthor{li2025preferenceleakagecontaminationproblem}
        }]
      ]
      [{Benchmark for detection methods}
        [{\citeauthor{shi2024detectingpretrainingdatalarge}, \citeauthor{duan2024membership}, \citeauthor{zhang-etal-2024-pretraining}, \citeauthor{ye-etal-2024-data}
        }]
      ]
    ]
    [How to \\ detect data \\ contamination \\ (Sec.~\ref{sec:how}), text width=4.7em
      [{White-Box\\Detection}
        [{\citeauthor{elangovan-etal-2021-memorization}, \citeauthor{brown2020language}, \citeauthor{touvron2023llama2}, \citeauthor{chowdhery2023palm}\\
        \citeauthor{achiam2023gpt}, \citeauthor{reimers2019sentence}, \citeauthor{lee2023platypus}, 
        \citeauthor{tu2024dicedetectingindistributioncontamination},
        \citeauthor{yang2023rethinkingbenchmarkcontaminationlanguage}
        \\
        }]
      ]
      [{Gray-Box\\Detection}
        [{\citeauthor{duan2024membership}, \citeauthor{ye-etal-2024-data}, \citeauthor{shi2024detectingpretrainingdatalarge}, \citeauthor{zhang2024minkimprovedbaselinedetecting}\\
        \citeauthor{zhang-etal-2024-pretraining}, \citeauthor{li2023estimatingcontaminationperplexityquantifying}, \citeauthor{wei2023skyworkopenbilingualfoundation},
        \citeauthor{zhang-etal-2024-pacost}
        }]
      ]
      [{Black-Box\\Detection}
        [{\citeauthor{golchin2023data}, \citeauthor{golchin2023time}, \citeauthor{duarte2024decopdetectingcopyrightedcontent}\\\citeauthor{dong-etal-2024-generalization},
        \citeauthor{ranaldi-etal-2024-investigating}, \citeauthor{chang-etal-2023-speak}, \citeauthor{deng2023investigating}, \citeauthor{carlini2021extracting}
        }]
       ]
    ]
    [{Future Directions\\(Sec.~\ref{sec:future directions})},
       [{(1)LLM Unlearning Methods (2)Robust Detection Methods\\(3)Distinguishing Between Data Contamination and Generalization
       }]
    ]   
]
\end{forest}
\caption{Structure of this paper}
\label{tab:tree}
\end{figure*}
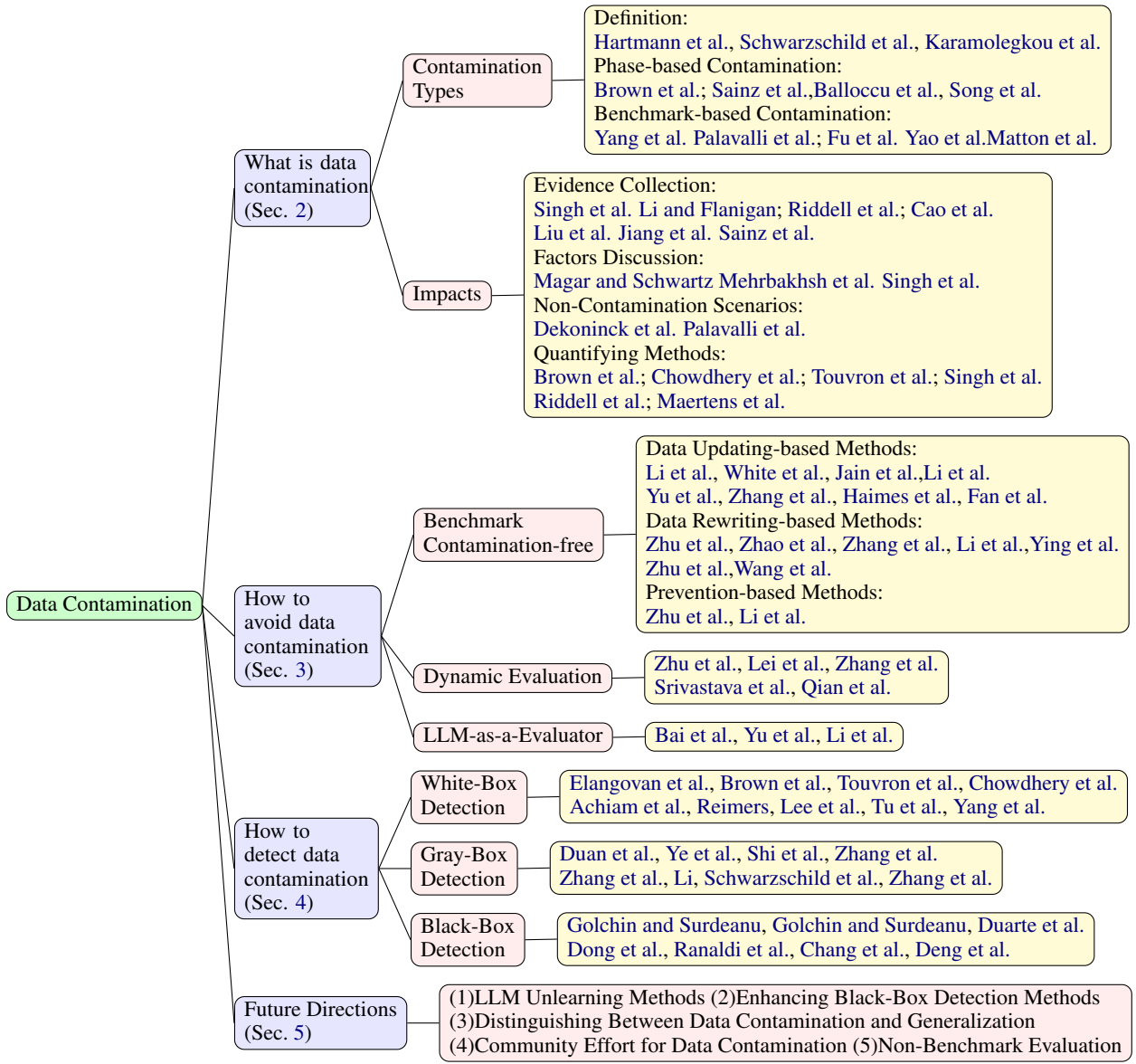

The organization of this paper is as follows, as shown in figure \ref{tab:tree}. In Section \ref{sec:what}, we discuss existing work on the definition and impacts of data contamination. Section \ref{sec:where}  summarizes current methods for constructing contamination-free datasets and dynamic evaluation approaches. Section \ref{sec:how} discusses how to detect data contamination. Finally, in Section \ref{sec:future directions}, we present several significant future challenges in this area.

\paragraph{Difference with previous survey}
Our paper systematically summarizes the definitions of data contamination across different scenarios and provides a fine-grained analysis of its impacts, particularly focusing on the characteristics of data contamination in Section~\ref{sec:characteristic}. Additionally, we enumerate several benchmarks utilized for quantifying data contamination in Section~\ref{sec:benchmark detection}.

\begin{table}[htbp]
\centering
\small
\begin{tabular}{|c|c|c|c|}
\hline
\textbf{Survey} & \textbf{Definition} & \textbf{Detection} & \textbf{Mitigation}\\ 
\hline
 \citeauthor{ravaut2024much}& $\times$  & $\checkmark$ & $\times$ \\
\hline
\citeauthor{xu2024benchmark} & $\checkmark$  & $\checkmark$ & $\checkmark$ \\
\hline
\citeauthor{fu2024does}  & $\times$  & Partial      & $\times$ \\
\hline
\citeauthor{chen2025recentadvanceslargelangauge} & $\times$  & $\times$     & Partial      \\
\hline
\citeauthor{deng-etal-2024-unveiling} & $\checkmark$  & $\checkmark$ & $\checkmark$ \\
\hline
Ours & Comprehensive  & $\checkmark$ & $\checkmark$ \\
\hline
\end{tabular}
\caption{Summary of Prior Surveys, $\checkmark$ means full coverage and $\times$ indicates it is not the main focus. Details are in Appendix \ref{sec:prior surveys}.}
\label{tab:survey-summary}
\end{table}

\section{What is Data Contamination}
\label{sec:what}
\subsection{Definition}
In recent years, a growing body of research has emerged to address the issue of data contamination in LLMs. However, the field lacks a standardized methodology to comprehensively summarize data contamination. Simply, let $\mathcal{D}_{\text{train}}$ denote the training dataset and $\mathcal{D}_{\text{test}}$ the evaluation dataset. Data contamination occurs when: $\mathcal{D}_{\text{train}} \cap \mathcal{D}_{\text{test}} \neq \emptyset$.
Building on this definition, our research extends the framework into two significant directions: (1) examining vulnerabilities across the entire lifecycle of LLMs, including pre-training, fine-tuning, and post-deployment contamination, and (2) addressing risks to benchmark integrity, such as data manipulation and potential label leakage.

\subsubsection{Phase-based Contamination}

\paragraph{Contamination during pre-training~\cite{sainz-etal-2023-nlp}} During initial training, web-scraped data often contains unwanted content (e.g., benchmark datasets like GLUE) because of imperfect filtering and deduplication (Lee et al., 2022). While complete prevention is impractical, transparency about pre-training data helps avoid biased evaluations (Dodge et al., 2021).

\paragraph{Contamination during fine-tuning~\cite{sainz-etal-2023-nlp}}
Directly fine-tuning models on benchmark data constitutes a significant form of data contamination. This practice is relatively uncommon in industrial settings~\cite{dekoninck2024evadingdatacontaminationdetection}, as developers prioritize maintaining model generalizability. However, in academic contexts, researchers deliberately employ this approach to create contaminated models for controlled experimental analysis.

\paragraph{Contamination after deployment~\cite{balloccu-etal-2024-leak}}
Post-deployment contamination introduces the notion of \textit{indirect data leakage}, where human interactions during model operation may inadvertently expose benchmark.

\paragraph{Multi-modal contamination~\cite{song2024textimagesleakedsystematic}}
Contamination in multi-modal models can occur in two primary ways. First, when text-label pairs or just the text inputs appear in the training corpus, creating direct overlap with test examples. Second, when triplets containing text, image, and labels appear in the training corpus, allowing models to memorize text-image-label triplet rather than learning generalizable capabilities.

\subsubsection{Benchmark-based Contamination}


\paragraph{Text contamination~\cite{jiang2024investigatingdatacontaminationpretraining}} Text contamination occurs when the input text components of evaluation samples appear in the pre-training corpus, creating an overlap between test and training data that may artificially inflate model performance metrics.

\paragraph{Text-label contamination~\cite{jiang2024investigatingdatacontaminationpretraining}} Text-label contamination refers to cases where the pre-training corpus contains not only input text but also prompts (task instructions) and corresponding labels or answers from evaluation samples, effectively exposing the model to both questions and correct answers prior to testing.

\paragraph{Augmentation-based contamination~\cite{palavalli-etal-2024-taxonomy}}
Augmentation-based contamination refers to data contamination caused by means such as sample masking (deleting key input/output content), noise injection (rewriting/replacing labels/alternative answers), or adversarial augmentation (adding distracting options/irrelevant context). 

\paragraph{Benchmark-level contamination~\cite{palavalli-etal-2024-taxonomy}}
Benchmark-level contamination arises when the model incorporates partial source corpora of benchmark datasets or outdated versions of these benchmarks during the training process.

\subsection{Impacts}
Data contamination critically undermines evaluation reliability and research validity. As~\cite{sainz-etal-2023-nlp,riddell2024quantifyingcontaminationevaluatingcode} demonstrated, benchmark overfitting can artificially inflate model performance and compromise scientific conclusions in NLP studies. For a comprehensive understanding of data contamination, we need to collect evidence, clarify non-contamination scenarios, quantify contamination and identify characteristics of data contamination.

\subsubsection{Evidence Collection}
\label{subsubsec:evidence}
Some papers reproduce results from previous work by attributing them to data contamination. \citet{li2024task} proposed evaluating models on training datasets with membership inference attacks, revealing contamination effects on performance. \citet{riddell2024quantifyingcontaminationevaluatingcode} quantitatively analyzes the overlap between popular code generation benchmarks and pretraining corpora and reveals data contamination. \citet{liu-etal-2024-evaluating} revealed that Chinese LLMs exhibit superficial knowledge despite extensive training, a phenomenon that may partially stem from data contamination. For a broad perspective, \citet{sainz-etal-2023-nlp} highlighted that current evidence on contamination remains fragmented across publications and informal channels, suggesting that the prevalence of contamination may be significantly underestimated.


\subsubsection{Non-Contamination Scenarios}
In this section, we explore non-contamination scenarios, where the overlap between training and testing data does not lead to performance improvement. \citet{dekoninck2024constat} established a causal relationship between model performance improvement and data contamination, explicitly defining cases where such overlap exists but does not enhance performance as non-contamination. Furthermore, \citet{palavalli-etal-2024-taxonomy} clarified several phenomena that improve performance on downstream tasks without being influenced by contamination. These include language understanding, prior task understanding, and transductive learning. These phenomena enhance empirical results while preserving the integrity of both the task and the model, distinguishing them from data contamination.

\subsubsection{Quantifying Contamination}
\label{chap:quantifying}

For model developers, the scenario for quantifying contamination focuses on using the N-grams algorithm to measure the overlap between the training and test data.
Contamination scoring mechanisms classify evaluation samples through threshold-based indices. 
For instance, \citet{brown2020language} used N-grams to evaluate contamination by checking whether each token in the tested sample appears in an n-gram from the pre-training corpus. 
In contrast, \citet{touvron2023llama2} introduced a method to align extensions between the testing samples and pre-training corpus, allowing mismatches in certain token positions using a "skip\_budget" hyperparameter. \citet{singh2024evaluation} further extended this method, focusing on the longest contaminated token span rather than all potential matches. 
In cases where this corpus is unavailable, search engines can be utilized to identify relevant problems and their corresponding solutions. If we can find it, it has likely already been used as training data by large language models~\cite{Li2023AnOS}.
Besides N-grams, \cite{cao2024concerneddatacontaminationassessing} proposed that current MIA-related metrics such as perplexity and Zlib compression entropy cannot effectively distinguish contaminated data from cleansed data, and there is a need for new metrics to quantify data contamination. \citet{singh2024evaluation} proposed a new contamination evaluation protocol, ConTAM, to explore how data contamination affects the evaluation results of LLMs, and provided a method to quantify it.


\subsubsection{Characteristics of Data Contamination}
\label{sec:characteristic}
In this section, we summarize five key characteristics of data contamination in large language models. 
\paragraph{The inevitability of data contamination}
As large language models (LLMs) continue to scale up, the size of their training datasets expands correspondingly~\cite{villalobos2024rundatalimitsllm}. These datasets are often sourced from extensive web crawls, which may inadvertently overlap with evaluation benchmarks, leading to data contamination~\cite{deng2023investigating}. This process is currently inevitable.
\paragraph{Scaling laws}
Larger models exhibit stronger contamination effects than smaller ones~\cite{kocyigit2025overestimationllmevaluationcontrolled}. As LLMs' memorization ability grows significantly with their model size, we argue it becomes even easier for them to reproduce training data instances~\cite{riddell2024quantifyingcontaminationevaluatingcode}. 
\paragraph{Cross-stage Characteristics}
Data contamination can occur during model pre-training, post-training, and deployment \cite{balloccu-etal-2024-leak}. \citet{kocyigit2025overestimationllmevaluationcontrolled} found that the training stage at which contamination occurs plays a crucial role in its impact. Early contamination leads to a sharp initial performance increase, but this effect gradually diminishes as training progresses. Late-stage contamination ultimately causes a larger performance gap. Uniform contamination (spread across the entire training process) produces the most lasting effects, with no significant spikes.
\paragraph{Task-specific Characteristics} 
Recent research indicates that data contamination impact varied across different tasks. Comparative experiments by \citet{palavalli-etal-2024-taxonomy} further reveal significant differences in the performance gains contributed by verbatim versus reformatted contamination in summarization tasks. Moreover, \citet{golchin2023time}, using an prompt-guided detection approach, demonstrated that the efficacy of the same contamination type varies markedly across different tasks during pretraining. \citet{jiang2024investigatingdatacontaminationpretraining} confirmed that the order and distribution of contaminated data moderate performance across different tasks, underscoring the need for evaluation protocols to be differentially designed for each task.

\paragraph{Cross-lingual Characteristics}
LLMs are overfitted to translated versions of benchmark test sets in non-English languages. This practice inflates model performance on the original English benchmarks without direct exposure to them~\cite{yao-etal-2024-data}, while evading existing detection methods~\cite{zhang2024careful}. \citet{kocyigit2025overestimationllmevaluationcontrolled} finds that contamination requires sufficient language representation to produce measurable effects: for resource-scarce languages, contamination has almost no impact on performance. These findings reveal the complexity and threshold effects of data contamination in multilingual environments.

\section{How to Avoid Data Contamination}
\label{sec:where}
This section discusses methods to avoid data contamination in evaluation. First, to reduce risks, benchmarks are often constructed following three strategies: data updating-based methods, data rewriting-based methods, and prevention-based methods. Second, dynamic evaluation generates adaptive samples using techniques like rule-based methods or agent-based methods. Finally, LLM-as-a-judge eliminates contamination risks, making it a key for contamination-free evaluation. However, there may be issues of preference contamination \cite{li2025preferenceleakagecontaminationproblem}.

\subsection{Benchmark Contamination-free Strategies}

Contamination-free benchmarking strategies ensure datasets stay up-to-date, preventing models from using outdated data. Data rewriting-based methods demonstrate the practical efficacy of paraphrasing techniques in contamination mitigation. Preventive measures involve technical defenses like encryption, access control, and de-contamination during inference to guarantee the reliability and fairness of LLM evaluation.

\subsubsection{Data Updating-based Methods}

Using the most recent data is intuitive for constructing contamination-free benchmarks, and some studies have proposed automatically collecting recent data to build questions. Meanwhile, recent data also need to maintain the stability of difficulty. 
LatestEval proposed an automated pipeline to dynamically generate contamination-free test sets from recent materials~\citep{li2024latestevaladdressingdatacontamination}. 
\citet{white2024livebench} introduced LiveBench, a dynamically updated benchmark that integrates tasks across math, coding, and reasoning with automated scoring to mitigate data contamination. Similarly, \citet{jain2024livecodebench} introduced LiveCodeBench, a code-generation benchmark that extends prior methodologies by dynamically evaluating self-repair capabilities and maintaining update cycles. 
\citet{fan2024nphardeval4vdynamicreasoningbenchmark} introduced NPHardEval4V-a dynamically updated benchmark to assess reasoning capabilities of MLLMs.
In code evaluation, EvoCodeBench~\cite{li2024evocodebench} is proposed to dynamically align with recent code repositories for fair evaluation.

\subsubsection{Data Rewriting-based Methods}

This type of methods use data augmentation to remove contamination from benchmarks, with LLMs' superior rephrasing and verifying capabilities. 
\citet{zhu-etal-2024-clean} proposed Clean-Eval to purify contaminated benchmarks by paraphrasing and back-translating data into semantically equivalent but lexically distinct forms.
\citet{zhao2024mmlu} proposed the MMLU-CF dataset, which is constructed by collecting diverse questions, cleaning data, sampling difficulty reasonably, checking data integrity with LLMs, and applying rewriting methods such as rephrasing questions and shuffling options to ensure the dataset remains contamination-free.
CLEVA~\citep{li-etal-2023-cleva} employs non-repetitive sampling and multi-strategy data rewriting for robust evaluation. \citet{ying2024automating} updated benchmarks with two strategies: style-preserving mimicry with LLMs and cognitive-level expansion using Bloom's taxonomy.  

\subsubsection{Prevention-based Methods}
Preventive measures focus on safeguarding test data integrity through technical and procedural controls. Core strategies include encrypting public test data with public-key cryptography, enforcing strict access permissions, and prohibiting derivative data creation. \citet{zhu2024inference} introduced Inference-Time Decontamination (ITD), a novel technique that identifies and rewrites potentially memorized responses during model inference. \citet{li2024c2levacomprehensivecontaminationfreelanguage} introduced C\textsuperscript{2}LEVA, a comprehensive bilingual benchmark with systematic contamination prevention mechanisms, which implements proactive measures such as test data rotation and enhanced encryption.

\subsection{Dynamic Evaluation}
\label{sec:dynamic evaluation}

\paragraph{Rule-based} Dynamic approaches address data contamination by leveraging adaptive assessment frameworks. \citet{zhu2024dyvaldynamicevaluationlarge} introduced DYVAL, a graph-based system that generates evaluation samples through algorithmic composition, constraint application, and functional descriptions. Its directed acyclic graph (DAG) architecture facilitates multi-step reasoning tasks with precisely controlled complexity. \citet{lei-etal-2024-s3eval} developed S3EVAL, a framework for SQL evaluation that utilizes randomized table-query pairs. This synthetic approach allows for customizable task lengths and difficulty levels, while systematically assessing long-context reasoning capabilities. \citet{zhang2024darg} proposed the DARG method, which dynamically generates evaluation samples with adjustable complexity and diversity using adaptive reasoning graphs. \citet{srivastava2024functional} introduced functionalization, a technique that transforms static question-answer pairs into parameterized code, enabling the generation of infinite test variants. \citet{qian2024varbench} further extended dynamic evaluation by perturbing key variables in questions, allowing for the dynamic generation of datasets with controlled variations.
\paragraph{Agent-based}
\citet{zhu2024dynamicevaluationlargelanguageMPA} proposed Multi-Principle Assessment (MPA), which utilizes LLM-based agents to automatically transform questions into new ones. \citet{wang2024benchmark} introduced a multi-agent framework to implement self-evolving benchmarks, which dynamically mutates question contexts and structures to update benchmarks.

\subsection{LLM-as-a-Judge}
\label{sec:llm driven evaluation}
Next-generation evaluation leverages LLMs themselves as assessment tools. They can serve the roles of scoring, ranking, and selection. \citet{bai2024benchmarking} presented the "LM-as-Examiner" framework, generating questions and evaluating responses through reference-free analysis. \citet{yu-etal-2024-kieval} deployed LLMs as "Interactors" in structured multi-turn dialogues that probe model capabilities while minimizing contamination risks.
\citet{li2024treeevalbenchmarkfreeevaluationlarge} proposed TreeEval, a benchmark-free system where LLMs generate hierarchical question trees. This adaptive approach adjusts difficulty based on model performance, creating unique assessment paths that prevent data contamination.

But \citet{li2025preferenceleakagecontaminationproblem} identified systematic bias in LLM-as-a-judge evaluations, where models trained on synthetic data from architecturally similar foundations receive unfair preference, compromising evaluation fairness.

\subsection{Benchmarks for Detection Methods}
\label{sec:benchmark detection}
In this section, we enumerate existing benchmarks utilized for quantifying data contamination. These datasets consistently comprise textual content paired with corresponding labels. The annotation of labels within these benchmarks is systematically conducted based on release dates, establishing a temporal framework for analysis.
WikiMIA\cite{shi2024detectingpretrainingdatalarge} datasets serve as a benchmark designed to evaluate membership inference attack (MIA) methods, specifically in detecting pretraining data from extensive large language models. BookMIA\cite{shi2024detectingpretrainingdatalarge}  utilizes book data to evaluate detection methods.
\citet{duan2024membership} introduce MIMIR, a Python package for evaluating memorization in LLMs, which presents greater challenges than WikiMIA. \citet{zhang-etal-2024-pretraining} introduce PatentMIA, specifically designed for Chinese-language pre-training data detection.
\citet{ye-etal-2024-data} propose a StackMIAsub dataset which supports most white- and black-box models, to evaluate detection methods.

\section{How to Detect Data Contamination}
\label{sec:how}

Data contamination detection involves identifying whether a text or dataset has been included in a model's training corpus. We categorize detection methods into three paradigms based on model access: white-box, gray-box, and black-box and analyse robustness in section \ref{sec:Robust Detection Methods}. For practical implementation guidance, we list some detection tools in Appendix~\ref{sec:Data contamination Detector}.

\subsection{White-Box Detection}
White-box methods use model internals or training data to detect contamination. 
\paragraph{N-gram based} The n-gram overlap method, known for its effectiveness and simplicity, is widely used for model developer contamination detection. Leading LLMs like LLaMA2 \cite{touvron2023llama2}, PaLM \cite{chowdhery2023palm}, and GPT-4 \cite{achiam2023gpt} stress the importance of detecting train/test overlaps.  
\citet{riddell2024quantifyingcontaminationevaluatingcode} quantifies the contamination levels between code generation benchmarks and pretraining corpora through surface-level and semantic-level methods involving n-gram matching.

\paragraph{Embeddings similarity based}
Embeddings similarity compares texts via cosine similarity of their embeddings (with a threshold value), capturing semantic relationships beyond lexical variations~\cite{reimers2019sentence}. \citet{lee2023platypus} used a similarity exclusion method based on embeddings, reducing dataset redundancy and filtering out duplicate data to ensure clean training data. To address sophisticated contamination forms, \citet{yang2023rethinkingbenchmarkcontaminationlanguage} introduced a hybrid approach combining embedding similarity search with GPT-4 powered semantic analysis. This detects paraphrased samples, enabling proactive benchmark decontamination.
\paragraph{Layer-specific} 
\citet{tu2024dicedetectingindistributioncontamination} proposed DICE to detect in-distribution contamination during fine-tuning by analyzing layer-specific activation patterns. This method trains contamination classifiers on sensitive intermediate layers, demonstrating a strong correlation between detection signals and performance inflation across multiple LLMs.

\subsection{Gray-Box Detection}
Gray-box approaches utilize partial model information, such as token probabilities, to compute perplexity or confidence, which can help detect data contamination. These methods mostly make a binary decision by comparing the score against a threshold value. \citet{duan2024membership} systematically investigated the underwhelming MIA performance on LLMs, identifying three primary contributing factors: the massive scale of training datasets that complicates memorization patterns, the limited number of training iterations that reduce model overfitting, and the inherently fuzzy decision boundaries between member and non-member samples. To address these shortcomings, the Min-K\% method established token-based effective methods using outlier token probabilities for pretraining data detection~\cite{shi2024detectingpretrainingdatalarge}. However, the effectiveness of this approach heavily depends on pre-designed K values and threshold parameters.
\citet{zhang2024minkimprovedbaselinedetecting} subsequently proposed Min-K\%++, theoretically grounding detection in local probability maxima identification and moved beyond heuristic-based methods. Similarly, based on the Min-K\% method, \citet{ye-etal-2024-data} introduced PAC, an MIA method that calculates polarization distances (based on Max $k_1$\% and Min $k_2$\%) through input perturbations. \citet{zhang-etal-2024-pretraining} proposed DC-PDD to employ corpus frequency divergences to reduce false positives. Without a threshold value, \citet{zhang-etal-2024-pacost} introduced PaCoST, a method that detects data contamination by statistically comparing confidence scores between original test items and their semantically equivalent paraphrased counterparts.
Alternative perplexity-based methods, \citet{li2023estimatingcontaminationperplexityquantifying} compare perplexity on benchmark samples against contaminated and clean baselines to show data contamination.Similar to this, \cite{wei2023skyworkopenbilingualfoundation} uses GPT-4 to generate data that is stylistically similar to the original GSM8K. The authors then compute the perplexity on the GSM8K training set (train), GSM8K test set (test), and GSM8K reference set (ref). 

\paragraph{Efficiency analysis} Token-based methods(Min-K\% \cite{shi2024detectingpretrainingdatalarge}, PAC \cite{ye-etal-2024-data}, Min-K\%++ \cite{zhang2024minkimprovedbaselinedetecting}, DC-PDD \cite{zhang-etal-2024-pretraining}) consist of a constant number of LLM forward passes and then some basic algebraic operations, making them efficient. And PaCoST \cite{zhang-etal-2024-pacost} requires computational resources for sample paraphrasing.

\subsection{Black-Box Detection}

Black-box methods operate without access to model internals, training corpus, primarily relies on the model's outputs for decision-making.
Specifically, these methods heavily rely on certain assumptions shown in Appendix~\ref{sec:definition of assumptions}.

\paragraph{Memorization-based methods}
Masked and completions are key practices in black-box methods.
~\citet{golchin2023time} first proposed a guided prompt-based detection method, which effectively identifies contamination in datasets through instance completion. \citet{golchin2023data} introduced DCQ, a multiple-choice question framework where each question presents an original instance alongside three perturbed versions (in which words are replaced with contextually relevant synonyms) and one invalid option. If the LLM consistently selects the original instance, this behavior may indicate data contamination. Similarly, \citet{chang-etal-2023-speak} introduced a challenging cloze task and used data archaeology to investigate the memorization of passages from 571 novels by LLMs. \citet{deng2023investigating} proposed TS-Guessing, a protocol designed to test a model's ability to reconstruct masked elements of test data. \citet{duarte2024decopdetectingcopyrightedcontent} developed DE-COP, a copyright detection framework that employs verbal versus paraphrased multiple-choice probing.  This approach uncovers subtle contamination in major benchmarks. As highlighted by \cite{ranaldi-etal-2024-investigating}, the Text-to-SQL task with GPT-3.5 involves data contamination, where the model is tasked with reconstructing masked column names using the table name, the remaining column names, and contextual information.

In contrast to previous studies, \citet{dong-etal-2024-generalization} introduced CDD to identify contamination by analyzing the peakedness of output distributions. When paired with the TED mitigation technique, the CDD approach effectively addresses both explicit and implicit forms of contamination while preserving the validity of model evaluations. Canary insertion \cite{carlini2021extractingtrainingdatalarge} involves retraining open models on synthetic benchmark-mimicking examples ("canaries") and measuring recall rates, with higher recall rates indicating a greater propensity for memorization.


\section{Future Directions}
\label{sec:future directions}
\subsection{LLM Unlearning Methods}
Unlearning techniques offer the potential to mitigate LLM privacy risks by erasing specific data elements. 
For language models, \cite{jang-etal-2023-knowledge} demonstrates that performing gradient ascent on target token sequences is an effective method for forgetting them. 
\citet{eldan2023whosharrypotterapproximate} is the first paper to present an effective technique for unlearning in large language models.
Future research should explore integrating contamination mitigation through unlearning methods. This emerging field shows promise and fundamental challenges. For instance, \citet{shumailov2024ununlearning} claims such data erasure may be fundamentally unachievable in current architecture.
\citet{shi2024detectingpretrainingdatalarge} uses Min-K\% to audit unlearning methods and they've found that some content still remains.

\subsection{Robust Detection Methods}
\label{sec:Robust Detection Methods}
Current detection methods confront several challenges.
Existing black-box contamination detection approaches rely on heuristic rules that \citet{fu2024does} showed fail under certain conditions, raising concerns about their fundamental reliability. Safety mechanisms within LLMs further complicate detection efforts, as methods that directly prompt for contaminated content often trigger filters that mask contamination indicators. Meanwhile, traditional methods may lack effectiveness in detecting 
augmentation-based contamination\cite{dekoninck2024evadingdatacontaminationdetection}. Future work should focus on developing detection methods with more robust assumptions that remain valid across a variety of scenarios and transformation strategies.


\subsection{Distinguishing Between Data Contamination and Generalization}
The ambiguity between contamination and generalization remains unresolved. 
\cite{Ishikawa_2025} systematically distinguishes data contamination from generalization in LLM benchmarks through a three-tier framework combining n-gram alignment, canary insertion, and perturbation testing. In this context, out-of-distribution (OOD) data performance is considered a true form of generalization. Building on this, future work should emphasize the distinctions between data contamination and generalization.


\section{Conclusion}

Our paper examines three fundamental perspectives in data contamination research: (1) defining data contamination through the lenses of phases and benchmarks; (2) exploring methodologies for conducting contamination-free evaluations, with a particular focus on dynamic evaluation and LLM-based assessment techniques; and (3) investigating methods for detecting data contamination, offering a comprehensive analysis of existing techniques and their limitations. 
Furthermore, we provide actionable recommendations for enhancing contamination-aware evaluation systems, aiming to foster more reliable LLM development practices.
\section{Limitations}
While we extensively cover various forms of data contamination, it is possible that new contamination mechanisms or models may not be fully captured in our analysis. Additionally, our focus is primarily on data contamination within the context of LLMs, and we may not have fully incorporated previous research on data contamination in other areas of machine learning. And there are so many static benchmarks that we only list some to demonstrate a contamination-free benchmark construction method. Additionally, as this survey focuses on LLM data contamination, we may not cover all related areas such as membership inference attacks (MIA), machine unlearning, and LLM memorization.


{
\bibstyle{acl_natbib}
\bibliography{custom,anthology}
\nocite{li2024awesome}
}

\appendix
\section{Data Contamination Evidence Collection Efforts}
\label{sec:Evidence Collection}
Several initiatives are currently collecting evidence on data contamination. Below are key platforms and resources involved in this effort:

\begin{itemize}
    \item \textbf{The Language Model Contamination Index (LM Contamination Index)}:  
    This is a database used to track and record evidence of language model contamination. For more information, visit: \url{https://hitz-zentroa.github.io/lm-contamination/}.
    
    \item \textbf{CONDA-Workshop Data Contamination Database}:  
    This is a community-driven project focused on the centralized collection of data contamination evidence. The goal is to help the community understand the extent of the problem and assist researchers in avoiding previous mistakes. Detailed information can be found at: \url{https://huggingface.co/spaces/CONDA-Workshop/Data-Contamination-Database}.
\end{itemize}

\section{Definition of Assumptions}
\label{sec:definition of assumptions}
\subsection{Verbatim Memorization}
\label{sec:verbatim memorization}
In the context of LLMs, verbatim memorization~\cite{carlini2021extracting,carlini2022quantifying} refers to the phenomenon where a model recalls exact sequences of text, often from the data it has been trained on. This occurs when a model has seen a specific passage or piece of information during its training process and is able to reproduce it exactly when prompted. Verbatim memorization can lead to issues of data contamination, where the model unintentionally outputs copyrighted or sensitive material verbatim, causing concerns regarding privacy, intellectual property, and validity in analytical tasks. 

\subsection{Black-Box Method Assumption}
\label{sec:assumption detail}
\citet{golchin2023data} has assumed that when a model has memorized instances from the original dataset, it will prefer selecting options containing the original instance over semantically similar perturbations. Additionally, LLMs may exhibit positional biases, where certain positions in multiple-choice options are more likely to be chosen, leading to potential overestimation or underestimation of contamination levels.

\citet {golchin2023time} gave the assumption that by providing a "guided instruction" with dataset name, partition information, and part of the reference instance, LLMs can generate the complete version of the data instance. This allows for calculating overlap between generated completions and reference instances, helping to infer whether the dataset partition is contaminated. 

\citet{duarte2024decopdetectingcopyrightedcontent} assumed that LLMs may memorize specific copyrighted content, such as books or academic papers, during training. When encountering similar content, they can distinguish whether they've seen it before. DE-COP exploits this by designing multiple-choice questions to test if the model can accurately identify original copyrighted content from paraphrased versions. Additionally, model selection biases can affect copyright detection results, and DE-COP introduces a calibration method to minimize such biases.

In \cite{dong-etal-2024-generalization}, it is assumed that contaminated training data significantly affects the output distribution of large language models. Specifically, when trained on contaminated data, the model's output distribution becomes more peaked, causing it to produce more consistent outputs on contaminated data, favoring outputs strongly correlated with the training data.

\citet{deng2023investigating} assumed that if an LLM can accurately guess missing parts of a test set, such as keywords or answer options, without external assistance, it suggests that the model has encountered the corresponding benchmark data during training. This indicates memorization-based contamination. The TS-Guessing protocol tests whether the model has memorized benchmark data by having it guess hidden information.

\citet{ranaldi-etal-2024-investigating} assumed that data contamination can be detected solely by analyzing the inputs and outputs of LLMs. For example, unusually high accuracy on tasks from datasets like Spider indicates that the model may have been exposed to this dataset during training, leading to memorization rather than genuine understanding. Additionally, data contamination may lead to inflated performance on zero-shot tasks when the model encounters potentially contaminated data during training.

\citet{chang-etal-2023-speak} assumed that LLMs may memorize portions of text from their training data, especially when evaluation datasets contain known texts. This memorization can lead to inflated performance on tasks such as code generation. Moreover, data repetition on the web—through search engines and open datasets—encourages memorization, which improves accuracy on tasks involving familiar content.

\subsection{Memorization and Data Contamination}

Instance-level contamination~\cite{fu2024does} does not always lead to verbatim memorization. Utilizing instance generation~\cite{carlini2022quantifying,karamolegkou-etal-2023-copyright}, demonstrates that verbatim memorization requires repeated exposures to this instance x during training. Indeed, future research on contamination should place more emphasis on LLMs' memorization. \citet{schwarzschild2024rethinkingllmmemorizationlens} proposed that strings can be considered memorized if they can be reproduced using a shorter prompt, while \citet{karamolegkou-etal-2023-copyright} investigated verbatim memorization, particularly in the context of copyrighted materials.

\section{Data Contamination Detector}
\label{sec:Data contamination Detector}
\citet{Li2023AnOS} present Contamination Detector to check whether test examples appear on the internet via Bing search and Common Crawl index. The tool is available at: \url{https://github.com/liyucheng09/Contamination_Detector}.

\citet{ravaut2024much} presented an open-source library for contamination detection in NLP datasets and LLMs. The library combines multiple methods for contamination detection and is available at: \url{https://github.com/liyucheng09/Contamination_Detector}.

Overlapy is a Python package developed to evaluate textual overlap (N-Grams) between two volumes of text. This tool can be accessed at: \url{https://github.com/nlx-group/overlapy}.

\citet{yao-etal-2024-data} introduced Deep Contam, a method that detects cross-lingual contamination, which inflates LLMs' benchmark performance while evading existing detection methods. An effective detection method is provided in the repository, accessible at: \url{https://github.com/ShangDataLab/Deep-Contam}.

\citet{tu2024dicedetectingindistributioncontamination} discussed the detection of in-distribution data contamination using LLM's internal state. The tool is available at: \url{https://github.com/THU-KEG/DICE}.

\citet{bordt2023testing, bordt2024colm} presented Tabmemcheck, an open-source Python library designed to test language models for memorization of tabular datasets. The package includes four different tests for verbatim memorization of a tabular dataset (header test, row completion test, feature completion test, first token test). It also provides additional heuristics to test what an LLM knows about a tabular dataset, such as feature names test, feature values test, dataset name test, and sampling. The package can be found at: \url{https://github.com/interpretml/LLM-Tabular-Memorization-Checker}.

\citet{yang2023rethinkingbenchmarkcontaminationlanguage} provided a package that includes the LLM decontaminator, which quantifies a dataset's rephrased samples relative to a benchmark. Based on the detection results, the contamination of rephrased samples in the dataset can be estimated and removed from the training set. This tool is available at: \url{https://github.com/lm-sys/llm-decontaminator}.

\section{Prior Surveys}
\label{sec:prior surveys}
Prior research on data contamination primarily focuses on three main areas: definition, detection, and mitigation. \citet{xu2024benchmark} and \citet{deng-etal-2024-unveiling} provide comprehensive surveys that thoroughly examine data contamination in large language models, covering conceptual definitions, detection methodologies, and mitigation strategies with similar classification frameworks for detection methods (matching-based and comparison-based). However, they differ significantly in their conceptualization of contamination types. The first paper primarily distinguishes between task-level contamination and language-level contamination, providing a function-oriented taxonomy. In contrast, the second paper presents a more granular severity-based hierarchy with four distinct levels: semantic-level (topical overlap), information-level (metadata and distributions), data-level (content without labels), and label-level contamination (complete exposure including ground truth). Their approaches to mitigation strategies also diverge notably. While the first paper emphasizes evaluation guidelines and procedural recommendations, the second paper offers a more structured framework categorized into three comprehensive strategies: data curation, data refactoring, and benchmark-free evaluation.

The remaining three studies focus on specialized subdomains of data contamination.
\cite{fu2024does} focuses on black-box detection assumptions and shows these methods fail under certain conditions, raising concerns about their fundamental reliability. \citet{ravaut2024much} investigates the critical issue of contamination in LLMs, categorizing it into data contamination and model contamination , while further distinguishing between input-only and input-label contamination scenarios. The authors systematically review state-of-the-art detection methods, including string matching, embedding similarity analysis, likelihood-based techniques, and novel LLM-driven approaches, highlighting their strengths and limitations. \citet{chen2025recentadvanceslargelangauge} conduct an in-depth analysis of existing static to dynamic benchmark aimed at reducing data contamination risks. Based on this, they propose a series of optimal design principles for dynamic benchmarking and analyze the limitations of existing dynamic benchmarks.

\end{document}